\documentclass[twocolumn]{article}
\usepackage{graphicx}
\usepackage{amsmath}
\usepackage{hyperref}

\title{All You Need is Sally-Anne: \\
ToM in AI Strongly Supported After Surpassing \\ Tests for 3-Year-Olds}

\author{
  Nitay Alon \\
  Department of Computer Science \\
  Hebrew University, Israel \\
  \texttt{Nitay.Alon@mail.huji.ac.il}
  \and
  Joseph Barnby \\
  Institute of Psychiatry, Psychology and Neuroscience, King's College London, UK \\
  Centre for AI and Machine Learning, Edith Cowan University, Western Australia, AU \\
  School of Psychiatry and Clinical Neuroscience, University of Western Australia, AU \\
  \texttt{Joe.Barnby@kcl.ac.uk}
  \and
  Reuth Mirsky \\
  Department of Computer Science \\
  Tufts University, MA, USA \\
  \texttt{Reuth.Mirsky@Tufts.edu}
  \and
  Stefan Sarkadi \\
  King's College London, UK \\
  Inria, France \\
  \texttt{Stefan.Sarkadi@kcl.ac.uk}
}

\date{}

\begin{document}

\maketitle

\begin{abstract}
Theory of Mind (ToM) is a hallmark of human cognition, allowing individuals to reason about others' beliefs and intentions. Engineers behind recent advances in Artificial Intelligence (AI) have claimed to demonstrate comparable capabilities. This paper presents a model that surpasses traditional ToM tests designed for 3-year-old children, providing strong support for the presence of ToM in AI systems.
\end{abstract}

\section{Introduction}
Theory of Mind (ToM) refers to the cognitive ability to understand that others have beliefs, desires, and intentions that may differ from one's own. It is a foundational component of social cognition, essential for predicting and interpreting others' behavior. 

A commonly used set of tasks to evaluate ToM capabilities is False-belief tasks. Two well-known examples include:

\begin{itemize}
    \item \textbf{Sally-Anne Test:} A task where a character named Sally places an object in one location and leaves. Another character, Anne, moves the object while Sally is away. The test evaluates whether participants understand that Sally will look for the object where she left it, despite its true location.
    \item \textbf{Smarties Task:} In this task, participants are shown a Smarties candy box that unexpectedly contains pencils. They are then asked what another person who has not seen the contents will think is inside. Success indicates an understanding that others can hold false beliefs.
\end{itemize}

\begin{figure}
    \centering
    \includegraphics[width=\linewidth]{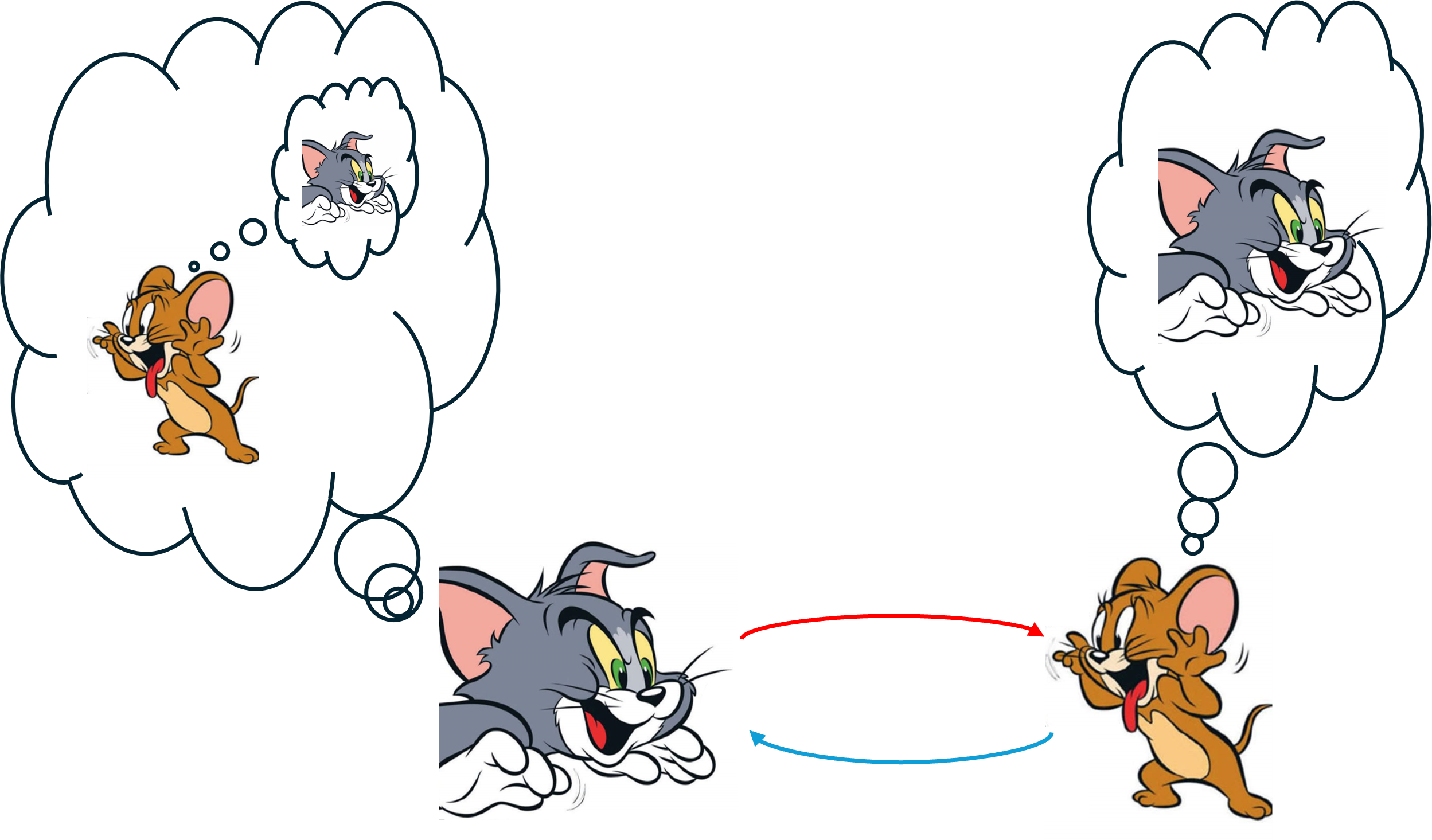}
    \caption{An illustration of a Theory of Mind process between two agents}
    \label{fig:enter-label}
\end{figure}

Attributing mental states to others is critical for collaborative and adaptive behaviours. While early AI systems lacked this capability, recent developments have introduced methods to endow agents with recursive reasoning skills. One primary way to achieve this has been to introduce gradient evaluation in an AI system. We introduce \textbf{ToM and GeRRI} --- ToM through Gradient Evaluation and Representation in Recursive Inference. ToM and GeRRI is an AI model that utilizes gradient evaluation for refining beliefs, represented using recursive inference. We show that this model was capable of passing the two aforementioned false-beliefs tests, showcasing a clear capability of ToM as human toddlers. This demonstrates that ToM may be achieved using a gradient evaluation algorithm.

\section{Related Work}
Theory of Mind has been extensively studied in both psychology and artificial intelligence. Early works such as \cite{premack1978does} laid the foundation for ToM in primates. \cite{baron1985does} introduced the Sally-Anne test to study ToM in children with Autism; a hypothesis was that those with Autism would not pass the test, and thus do not possess ToM. In AI, \cite{rabinowitz2018machine} demonstrated machine ToM using meta-learning. \cite{zhu2021recursive} explored recursive neural networks for ToM modelling. Other notable contributions include \cite{baker2017rational} on Bayesian mechanisms underlying ToM, \cite{gopnik1992child} on developmental trajectories of ToM, and \cite{deweerd2013recursive} on recursive reasoning.

Recent works such as ToMnet \cite{rabinowitz2018machine} and BERT-based ToM models \cite{chen2022bert} show the evolution of ToM in AI. \cite{tomasello2005understanding} emphasized shared intentionality, while \cite{frith2006social} focused on social neuroscience aspects. These studies collectively highlight the importance of recursive inference and gradient-based learning in achieving ToM in AI.

\section{ToM and GeRRI}
Our approach leverages gradient-based evaluation techniques to capture recursive belief structures. The key technical contributions include:

\paragraph{Gradient-Based Inference:} Gradient-based inference (GBI) serves as the backbone of our method, leveraging backpropagation to iteratively adjust belief representations. This mechanism allows the model to fine-tune its understanding of others' latent mental states by minimizing prediction errors through continuous feedback. Bayesian inference is central to our approach, providing a formal probabilistic structured framework for updating beliefs based on observed data. The fundamental equation is given by:

\begin{equation}
P(H | D) = \frac{P(D | H) \cdot P(H)}{P(D)},
\end{equation}

where $H$ denotes the hypothesis regarding a teammate's mental state, and $D$ represents observed actions. This formulation enables the model to incorporate prior knowledge while dynamically adjusting beliefs based on new observations. Minimizing the prediction error between expected and observed behaviour using GBI improves ToM predictability in test time. 

Moreover, our model integrates stochastic gradient descent (SGD) with adaptive learning rates to enhance convergence speed and stability. This flexibility allows the system to cope with non-stationary environments where teammates' behaviours may evolve.

\paragraph{Recursive Representation:} Human cognition can engage in recursive reasoning, contemplating what others think about one's thoughts. Our model simulates this capacity by nesting inferences to achieve multiple layers of reasoning. Recursive inference is formalized as:

\begin{equation}
P(H_k | D_t) \propto \sum_{k} P(d_t | D_{t-1}, H_{k-1}) \cdot P(H_{k} | D_{t-1}),
\end{equation}

where $H_{k}$ represents a higher-order belief and $H_{k-1}$ a lower-order belief and $d_t$ is the observed behaviour at time $t$. This hierarchical structure enables deeper reasoning beyond first-order beliefs, reflecting a sophisticated understanding of teammates' intentions. These nested beliefs may also account for the beliefs of the acting agent about the beliefs other ascribe to them, allowing our model to answer questions such as:"If you know that the Smarties can is full of pencils, and you know that Adam doesn't know this --- what would Adam say that you'll think about the content of the can"?

The recursive structure is implemented through a recurrent neural network (RNN) architecture with gated mechanisms to manage information flow. By retaining relevant historical information while discarding noise, the RNN captures long-term dependencies critical for complex multi-agent interactions. Using GBI we regularize the network's deviations from the expected outcome.
Empirical results demonstrate that deeper recursion levels correlate with improved predictive accuracy, indicating that multi-layered reasoning is essential for nuanced social cognition.

\paragraph{Representation Learning:} Effective structured representation of mental states is crucial for successful inference. Our neural architecture dynamically encodes these representations through continuous gradient updates, allowing the model to adapt to varying contexts and teammate behaviours. The training objective is designed to minimize a composite loss function that includes prediction error and a regularization term to ensure consistency with known distributions.

A key component of our objective function is the Kullback-Leibler (KL) divergence, which measures the discrepancy between the inferred posterior distribution and an approximated distribution $Q(H)$. The loss function is defined as:

\begin{equation}
\mathcal{L} = \text{KL}(P(H | D) || Q(H)),
\end{equation}

Minimizing this loss encourages the model to produce accurate and well-calibrated belief distributions. In addition, we incorporate an entropy regularization term to maintain exploration during early training stages, preventing the model from converging prematurely to suboptimal belief representations.

Finally, our approach benefits from multi-task learning, where the same representation is utilized for multiple inference tasks, promoting knowledge transfer and reducing overfitting. The synergy of gradient-based optimization, recursive reasoning, and dynamic representation learning forms a comprehensive framework for achieving Theory of Mind in AI systems.

\section{Experimental Results}

\begin{figure}
    \centering
    \includegraphics[width=\linewidth]{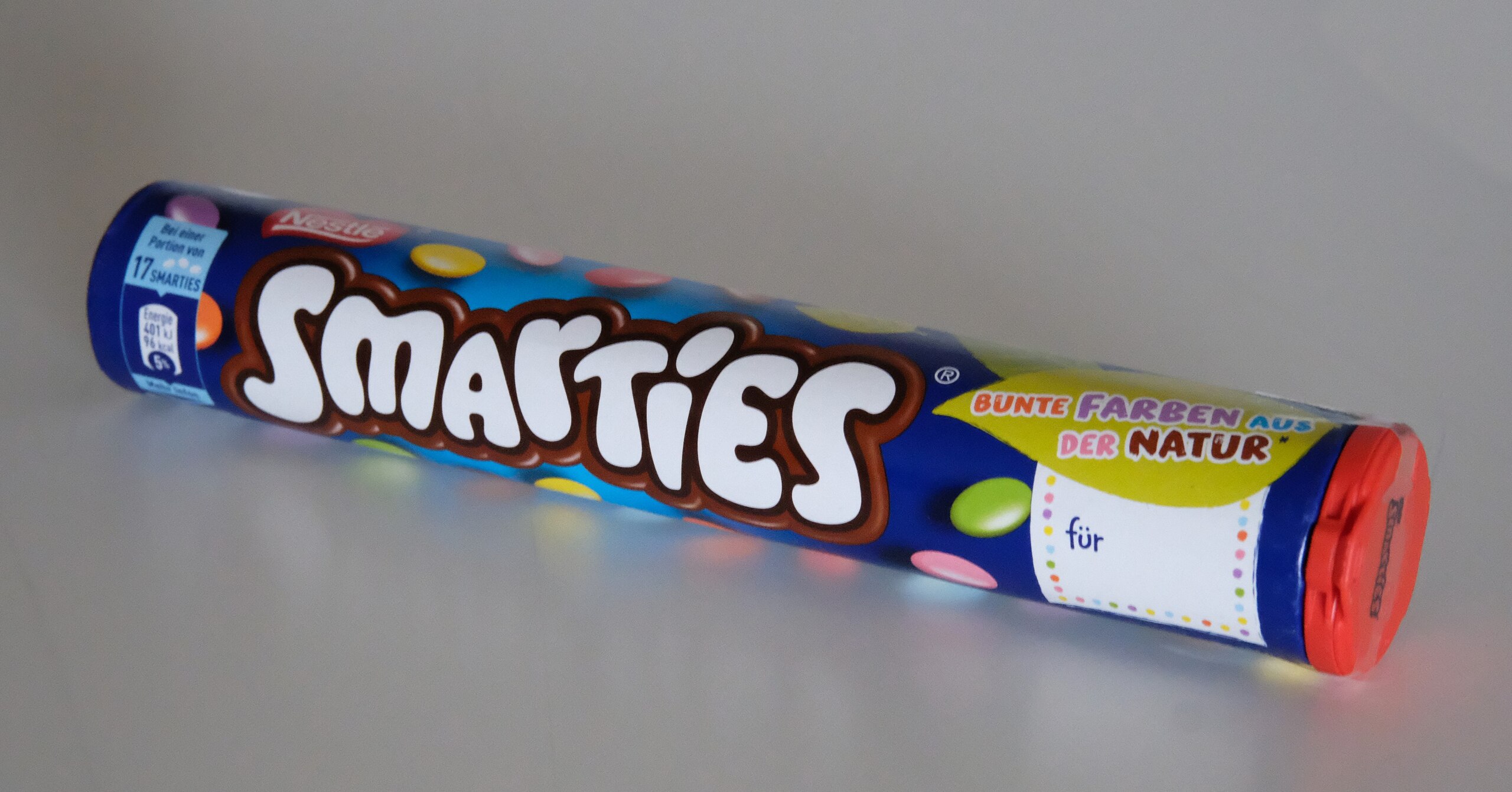}
    \caption{A can of Smarties full of pencils.}
    \label{fig:enter-label}
\end{figure}

We conducted two primary trials to evaluate our model's capacity for Theory of Mind (ToM): the Sally-Anne task and the Smarties task. Both are well-established false-belief tests commonly used to assess ToM in young children.

\paragraph{Sally-Anne Task:} In this scenario, the model must predict where Sally will look for the object. Our model successfully inferred Sally's outdated belief, performing on par with the average accuracy of 3-year-old toddlers as reported in psychological studies.

\paragraph{Smarties Task:} In this task, the model is asked to predict what a new, uninformed agent would believe is inside the Smarties tube (Figure 2). Similar to the Sally-Anne task, our model demonstrated performance comparable to 3-year-old toddlers, correctly attributing a false belief to the uninformed agent in the majority of trials.

\paragraph{Experimental Setup and Procedure:} For both tasks, we trained the model on a dataset comprising simulated agent interactions designed to emulate real-world false-belief scenarios. The training regimen emphasized recursive inference and adaptive belief updating. We evaluated the model across 100 independent trials for each task, measuring its ability to generate accurate belief predictions under conditions of partial and conflicting information.

\paragraph{Results:} Our findings show that the model consistently achieved performance levels similar to those of 3-year-old children in classic ToM tasks. The results support our hypothesis that gradient-based recursive reasoning can replicate early-stage human social cognition. The use of a Smarties-like tube with pencils (Figure 2) effectively demonstrated the model's capability to handle counterfactual reasoning, a crucial aspect of ToM.

\section{Conclusion}

In conclusion, our April Fool's submission intended to \textbf{highlight critical gaps in research methodology of Theory of Mind (ToM) in AI compared to cognitive sciences}:

First, while false-belief tasks like the Sally-Anne and Smarties tests are iconic, they represent only a narrow slice of ToM capabilities, often calibrated in very specific experimental settings. Real-world social cognition involves a broader spectrum of abilities, including self-other representation, metacognition, and generalization \cite{barnby2024beyond}, and should be able to complete these tasks with little conceptual sophistication \cite{butterfill2013construct}. 

Evaluating large language model (LLM)-based algorithms using existing ToM tests is problematic. These tests were originally designed to assess human cognitive development to diagnose Autism, and may not translate meaningfully to artificial systems. In humans, it has been argued that the best way to capture multiple facets of ToM is through live, dynamic social scenarios \cite{rusch2020theory, barnby2023formalising} rather than static vignettes, and thus to interpret an LLMs performance, computer science should follow suit. LLMs often excel at pattern recognition, resulting in high performance that may be mistaken for true, structured, model-based understanding that can generalize, creating a Clever Hans effect \cite{shapira2024clever}.

Additionally, using gradient-based methods for recursive reasoning is inherently problematic --- not completely disregarding its potential, but it should be used carefully. Gradient descent is fundamentally designed for optimization rather than hierarchical belief modelling. Recursive inference demands non-stationary representations across multiple levels of abstraction, which gradient-based approaches struggle to maintain. This causes inherent instability or convergence to trivial solutions. That said, a lot of successful research started with someone saying it could not be done.

We recommend that researchers critically reassess current methodologies and explore alternative frameworks better suited for AI systems. For more insights and ongoing discussions, we encourage readers to visit the website of the Theory of Mind for AI (ToM4AI) initiative\footnote{\url{https://sites.google.com/view/theory-of-mind-aaai-2025/main}}, which held its first workshop at AAAI 2025.

\bibliographystyle{plain}
\bibliography{references}

\end{document}